\documentclass{article}

\usepackage[preprint]{neurips_2024}

\usepackage[utf8]{inputenc} %
\usepackage[T1]{fontenc}    %
\usepackage{hyperref}       %
\usepackage{url}            %
\usepackage{booktabs}       %
\usepackage{amsfonts}       %
\usepackage{nicefrac}       %
\usepackage{microtype}      %
\usepackage{xcolor}         %
\usepackage{color}
\usepackage[utf8]{inputenc}
\usepackage{longtable}
\usepackage{graphicx}
\usepackage{threeparttable}
\usepackage{xspace}
\usepackage{appendix}
\usepackage{booktabs}
\usepackage[utf8]{inputenc}
\usepackage{textcomp}
\usepackage{amsmath, amssymb, amsthm}
\usepackage{enumitem}
\usepackage{algorithm}
\usepackage[noend]{algpseudocode}
\algtext*{EndProcedure}%
\usepackage{xcolor,colortbl}
\usepackage{footnote}
\usepackage{url}
\usepackage{tabularx}
\usepackage{multicol}
\usepackage{multirow}
\usepackage{pifont}
\usepackage{bm}
\usepackage{listings}
\usepackage{relsize}
\usepackage{mathtools}
\usepackage{wrapfig}
\usepackage{bbm}
\usepackage{blindtext}
\usepackage{bm}
\usepackage{balance}
\usepackage{mdwlist}
\usepackage{wrapfig}
\usepackage{caption}
\usepackage{subcaption}
\usepackage{makecell}
\usepackage{appendix}
\usepackage{graphicx}

\setlist[enumerate]{leftmargin=*}
\setlist[itemize]{leftmargin=*}

\usepackage[most]{tcolorbox}
\newtcolorbox{mybox}{
  colback=gray!10!white,
  colframe=black,
  rounded corners,
  boxrule=1pt,
  boxsep=5pt,
  left=2pt, right=2pt, top=2pt, bottom=2pt
}
\definecolor{deepgreen}{rgb}{0.0, 0.55, 0.0}
\definecolor{deepred}{rgb}{0.65, 0, 0.0}

\DeclareRobustCommand*{\RaiseBoxByDepth}{%
    \raisebox{-0.2\height}%
}

\usepackage{wasysym}
\usepackage{amssymb}%
\usepackage{pifont}%
\usepackage{mdframed}
\usepackage{setspace}

\newcommand{\method}{\textsc{LinkGPT}\xspace}

\newcounter{theorem}

\newcommand{\mypar}[1]{\vspace{0.5mm}\noindent\textbf{#1}}

\title{\method: Teaching Large Language Models To Predict Missing Links}

\author{Zhongmou He, Jing Zhu, Shengyi Qian, Joyce Chai, Danai Koutra\\
  University of Michigan\\
}

\begin{document}

\maketitle

\begin{abstract}
Large Language Models (LLMs) have shown promising results on various language and vision tasks. Recently, there has been growing interest in applying LLMs to graph-based tasks, particularly on Text-Attributed Graphs (TAGs). However, most studies have focused on node classification, while the use of LLMs for link prediction (LP) remains understudied. In this work, we propose a new task on LLMs, where the objective is to leverage LLMs to predict missing links between nodes in a graph. This task evaluates an LLM's ability to reason over structured data and infer new facts based on learned patterns. This new task poses two key challenges: (1)~How to effectively integrate pairwise structural information into the LLMs, which is known to be crucial for LP performance, and (2)~how to solve the computational bottleneck when teaching LLMs to perform LP. To address these challenges, we propose \method, the first end-to-end trained LLM for LP tasks. To effectively enhance the LLM's ability to understand the underlying structure, we design a two-stage instruction tuning approach where the first stage fine-tunes the pairwise encoder, projector, and node projector, and the second stage further fine-tunes the LLMs to predict links. To address the efficiency challenges at inference time, we introduce a retrieval-reranking scheme. Experiments show that \method can achieve state-of-the-art performance on real-world graphs as well as superior generalization in zero-shot and few-shot learning, surpassing existing benchmarks. At inference time, it can achieve $10\times$ speedup while maintaining high LP accuracy. \end{abstract}

\section{Introduction}
\label{sec:intro}
Graph-structured data is ubiquitous in real-world applications, ranging from social networks and recommendation systems to biological networks and knowledge graphs. Among the various tasks performed on graphs, link prediction (LP) is of particular importance, as it enables the discovery of missing or potential connections between nodes. Studies have also found that the performance of the link prediction task is directly related to the model's ability to understand the underlying structure~\cite{Mao2023RevisitingLP, zhu2023touchup, shomer2023adaptive}. %
In recent years, LP has been tackled using Graph Neural Networks (GNNs), which learn node representations by aggregating information from their local neighborhoods~\cite{zhang2018link, zhang2021labeling, liben2003link}. However, GNNs often struggle to capture long-range dependencies and complex semantic information present in text-attributed graphs (TAGs) and fail to generalize to unseen graphs ~\cite{jin2023patton}.

Recent advancements in Large Language Models (LLMs) have revolutionized natural language processing (NLP) and computer vision tasks~\cite{alayrac2022flamingo, li2023blip2, touvron2023llama, liu2023llava}. LLMs have demonstrated remarkable capabilities in understanding and generating human-like text, thanks to their ability to capture rich semantic information from large-scale pretraining~\cite{openai2023gpt4}. Inspired by their success, researchers have begun exploring the application of LLMs to graph-based tasks. However, most existing works focus on node classification which primarily involves node-level information.
It remains unclear whether and how LLM can be used for LP, a task that relies on pairwise information~\cite{tang2023graphgpt,he2023harnessing}.

Therefore, we propose to explore the ability of LLMs to reason over the graph structure for the task of link prediction, where the objective is for LLMs to predict missing links between nodes in a graph, given the existing graph structure and node attributes (i.e., link prediction
in text-attributed graphs). 

Nonetheless, teaching LLMs to perform LP is not straightforward and presents two key challenges. First, LLMs are primarily designed to process textual and sequential data and how to integrate node-wise and pairwise structural information into LLMs is non-trivial. Second, teaching LLMs to predict missing links presents computational bottlenecks. LP typically requires ranking against a large number of negative candidates (e.g., 1,000) for each link at inference time, which is computationally expensive when combined with LLMs. Efficiently and accurately ranking candidates is crucial for the practicality of LLM-based LP methods. 

To address these challenges, we propose \method, the first end-to-end trained LLM specifically designed to predict missing links in TAGs. \method consists of three main components: (1)~node encoding and pairwise encoding, (2)~two-stage instruction tuning, and (3)~a retrieval-reranking scheme for inference. Experiments show that \method not only achieves state-of-the-art performance across datasets but also has strong generalization ability in both  
zero-shot and few-shot in-context learning settings. When employing our designed retrieval-rerank scheme for inference, \method is $10\times$ faster while maintaining high LP accuracy. %

Our contributions are summarized as follows:
\begin{itemize}

\item \textbf{First LLM designed for LP:} We introduce the first-of-its-kind link prediction approach that teaches LLMs to predict missing links in graphs, enabling LLMs to reason over structured data.

\item \textbf{Effective and Efficient Method:} We propose \method, a framework that effectively incorporates node-wise and pairwise structural knowledge into LLMs via instruction tuning and significantly improves inference time efficiency by the retrieval-rank scheme. 

\item \textbf{Extensive Evaluation:} We conduct extensive experiments on real-world TAGs, demonstrating \method's superior performance and fast inference compared to existing state-of-the-art methods. Moreover, we showcase LinkGPT's ability of zero-shot generalization and few-shot in-context learning to cross-category and cross-domain unseen datasets. 
\end{itemize} 
\section{Related Work}
\label{sec:related}
\noindent \textbf{Link prediction.} Link prediction aims to complete missing links in a graph, with applications ranging from knowledge graph completion to e-commerce recommendations ~\cite{martinez2016survey, vashishth2019composition}. While heuristic algorithms were once predominant, Graph Neural Networks (GNNs) for Link Prediction have been prevalent in the past few years. Methods that use GNNs for LP mainly fall into two categories: Graph Autoencoder (GAE)-based methods and enclosing subgraph-based methods~\cite{kipf2016variational, chamberlain2023graph, zhang2018link}. While GNN-based methods show promising results, they do not generalize to unseen graphs. In this work, we aim to explore LLMs' structure reasoning ability on LP tasks and evaluate their generalization ability in both zero-shot and few-shot settings.

\vspace{0.25em}
\noindent \textbf{Large Language Models for graphs.}
Recent progress on applying LLMs for TAGs aims to leverage the power of LLMs to boost performance on graph-related tasks ~\cite{he2023harnessing, jin2023large}. Two main strategies have emerged: (1) LLMs as predictors, where LLMs directly generate solutions to tasks like node classification and link prediction ~\cite{tang2023graphgpt, chen2024llaga}, and (2) LLMs as enhancers,  which utilize LLM capabilities to improve the representations learned by smaller GNNs for better efficiency ~\cite{zhu2024efficient}. \method directly predicts missing links in graphs and thus falls into the first category. While previous works mainly focus on improving LLMs' ability to predict node labels, here we believe that LLMs' ability to predict missing links in graphs presents its actual structure reasoning ability.

\vspace{0.25em}
\noindent \textbf{Instruction tuning.}
Instruction tuning has emerged as a promising approach to adapting large language models (LLMs) for various tasks and modalities. Several recent works have explored instruction tuning in the context of vision-language models and multimodal learning ~\cite{liu2023llava, li2023blip2, zhu2023minigpt}. Instruction tuning has also been employed in several LLM-based models for graph tasks~\cite{tang2023graphgpt, chen2024llaga}. Our work further explores instruction tuning in the context of structure-semantic understanding. 
\section{Preliminaries}
\label{sec:preliminaries}

\noindent \textbf{Graphs.}
We consider a \textbf{graph} $\mathcal{G}=\{\mathcal{V}, \mathcal{E}, \mathcal{X}\}$, where $\mathcal{V}$ is the set of vertices, $\mathcal{E}$ is the set of edges, and $\mathcal{X}$ denotes the set of attributes or features associated with each node in $\mathcal{V}$. A graph $\mathcal{G}$ is considered to be a \textbf{Text-Attributed Graph (TAG)} if its node attributes $\mathcal{X}$ are in textual format, i.e. each node has text features. We further denote $\mathcal{N}_u^k$ as the \textbf{$k$-hop neighbors} of node $u$, i.e., the set of nodes at a distance equal to $k$ from $u$. 

\vspace{0.05cm}
\noindent \textbf{Link prediction.} Given a source node $s\in\mathcal{V}$ and a set of candidate target nodes $\mathcal{C}=\{t_1,t_2,\cdots,t_{N_C}\}$, which consists of  $1$ positive target node and $N_C-1$ negative target node, the link prediction task aims to rank all candidate target nodes based on the probability that there is a link between $s$ and $t_i$.

\section{Method}
\label{sec:method}

\begin{figure}
    \centering
    \includegraphics[width=1\linewidth]{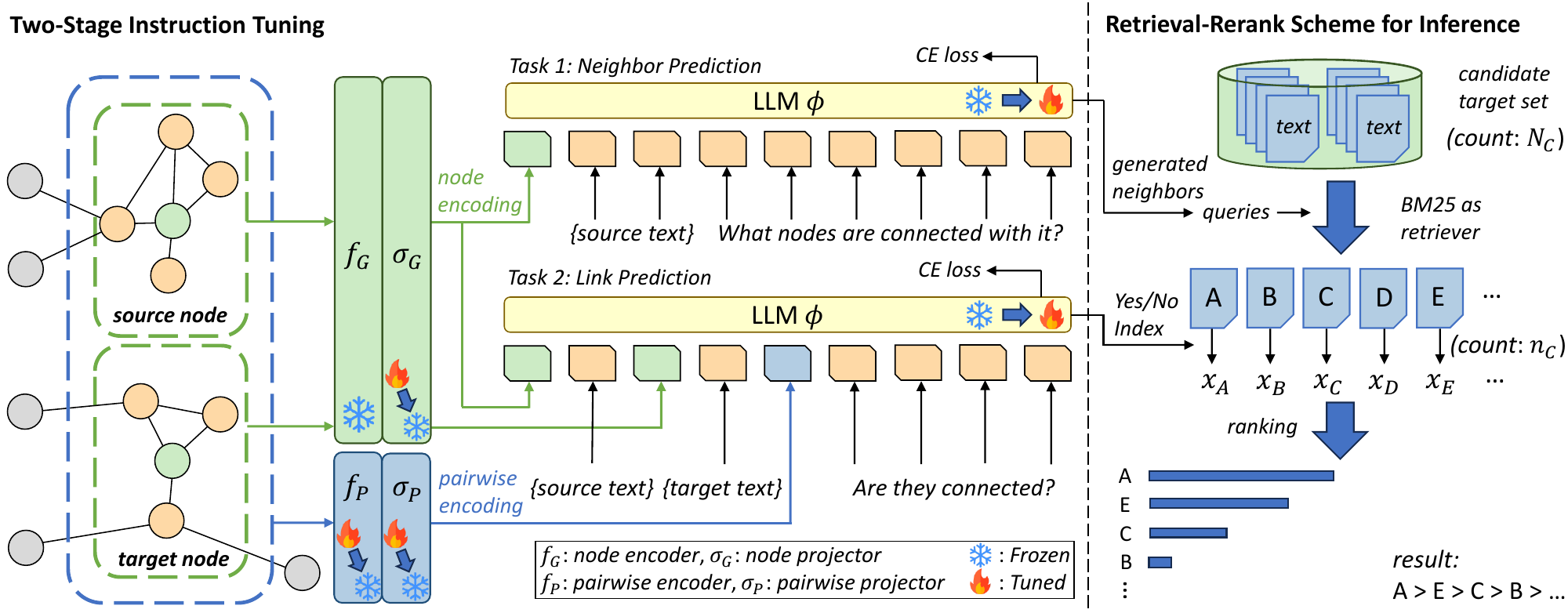}
    \caption{Overview of the \method framework. The framework consists of three main components: (1) node encoding and pairwise encoding, (2) two-stage instruction tuning, and (3) a retrieval-rerank scheme for inference. In this figure, \RaiseBoxByDepth{\includegraphics[height=1em]{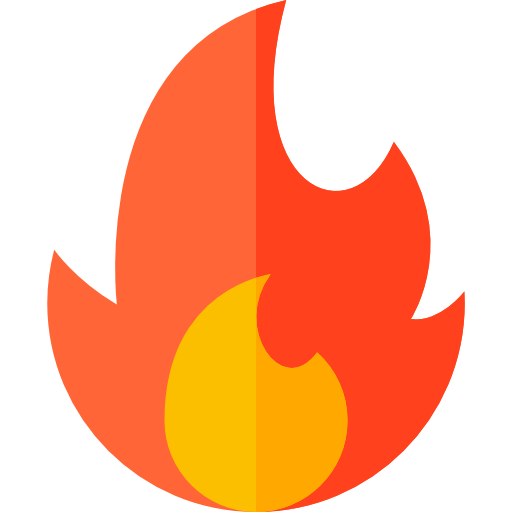}} $\Rightarrow$ \RaiseBoxByDepth{\includegraphics[height=1em]{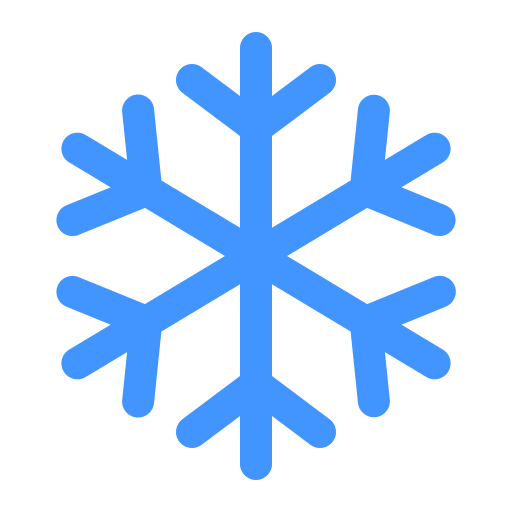}} means that the corresponding module is tuned during stage 1 and is frozen during stage 2, and vice versa. Note that the two LLMs in this figure are the same model.}
    \label{fig:architecture}
\end{figure}

In this section, we present the proposed \method framework for link prediction in text-attributed graphs (TAGs). \method leverages the power of Large Language Models (LLMs) to effectively capture both graph structure and textual semantics, enabling accurate and efficient link prediction. The framework consists of three main components: (1) node encoding and pairwise encoding, (2) two-stage instruction tuning, and (3) a retrieval-rerank scheme for inference. An overview of the \method framework is illustrated in Figure \ref{fig:architecture}.

\subsection{Node Encoding and Pairwise Encoding}

To effectively capture graph structure and textual semantics, we first introduce two special tokens: \texttt{<NODE>} and \texttt{<PAIRWISE>}. The \texttt{<NODE>} token represents a node and its neighborhood, while the \texttt{<PAIRWISE>} token encodes the pairwise relationship between two nodes.
 
\mypar{Neighborhood-aware node encoding.}
We obtain neighborhood-aware node embeddings through contrastive graph-text pre-training (CGTP). Let $f_G$ and $f_T$ denote the node encoder and text encoder, respectively. The node representations $\mathbf{H}\in \mathbb{R}^{N\times d_H}$ and neighborhood text representations $\mathbf{T}\in\mathbb{R}^{N\times 2d_T}$ are obtained as follows:
\begin{equation}
    \mathbf{T'}=f_T(\mathcal{X}), \quad \mathbf{T}_{v}=\text{concat}\left(\mathbf{T}_{v}', \frac{1}{|\mathcal{N}_v^1|}\sum\nolimits_{u\in\mathcal{N}_v^1}\mathbf{T}_{u}'\right), \quad \mathbf{H}=f_G(\mathcal{V}, \mathcal{E}, \mathcal{X})
\end{equation}
where $\mathbf{T}_{v}'\in \mathbb{R}^{d_T}$ is the raw text encoding of node $v$. We employ GraphFormers~\cite{Yang2021GraphFormersGT} as our node encoder $f_G$, which fuses text encoding and graph aggregation into a uniform iterative workflow, and we use BERT~\cite{devlin2018bert} as our text encoder $f_T$. The contrastive loss $\mathcal{L}$ is calculated as follows:
\begin{equation}
    \Gamma=\hat{\mathbf{H}}\hat{\mathbf{T}}^T / \tau, \quad \mathcal{L}=((\text{CE}(\Gamma, \mathbf{y})+\text{CE}(\Gamma^T, \mathbf{y}))/2
\end{equation}
where $\hat{\mathbf{T}}=\text{norm}(\mathbf{T})$, $\hat{\mathbf{H}}=\text{norm}(\mathbf{H})$, $\tau$ denotes the temperature, and the label $\mathbf{y}$ is set to $(0,1,\cdots,N-1)^T$. 
The learned node encoding $\mathbf{H}$ is then mapped to the semantic space of the LLM using a simple alignment projector $\sigma_{\text{G}}$. Details about node encoding are in Appendix \ref{sec:graphformers}.

\mypar{Pairwise encoding.} 
To capture the pairwise relationship between nodes, we employ LPFormer \cite{shomer2023adaptive}. 
For node $a$ and $b$, the LPFormer learns a pairwise encoding $f_P(a, b)$ to represent their relationship.
\begin{equation}
    f_P(a,b)=\sum\nolimits_{u\in\mathcal{V}_{(a,b)}}w(a,b,u)\odot h(a,b,u)
\end{equation}
where $w(a,b,u)$ and $h(a,b,u)$ denote the importance and encoding of node $u$ relative to the relation between $a$ and $b$ respectively, and $\mathcal{V}_{(a,b)}$ denotes the set of nodes that might be important to the relation. $f_P(a,b)$ is also mapped to the semantic space of LLM using an alignment projector $\sigma_{\text{P}}$. Note that the pairwise encoder is not pre-trained. Details about pairwise encoder are in Appendix \ref{sec:pairwise}.

\subsection{Two-Stage Instruction Tuning}

\begin{figure}[t]
\centering
\begin{mybox}
\textbf{Link Prediction:}

This is the source node. \texttt{<NODE>} Text: \textit{\{text\}}.

This is another node. \texttt{<NODE>} \texttt{<PAIRWISE>} Text: \textit{\{text\}}. Is this node connected with the source node? Answer: \textcolor{deepgreen}{\textit{\{Yes/No\}}}

This is another node. \texttt{<NODE>} \texttt{<PAIRWISE>} Text: \textit{\{text\}}. Is this node connected with the source node? Answer: \textcolor{deepgreen}{\textit{\{Yes/No\}}}

...

\textbf{Neighbor Prediction:}

This is the source node. \texttt{<NODE>} Text: \textit{\{text\}}. What nodes are connected with it? 

Answer: Text: \textcolor{deepgreen}{\textit{\{text\}}}. Text: \textcolor{deepgreen}{\textit{\{text\}}}. ...
\end{mybox}
\caption{Instruction templates for link prediction and neighbor prediction tasks. Only the answers (highlighted in green) are used for calculating the loss.}
\label{fig:instruction}
\end{figure}

We leverage a two-stage instruction tuning approach to train \method for link prediction and neighbor prediction tasks. The instruction templates for these tasks are shown in Figure \ref{fig:instruction}.

\mypar{Prompt design.} 
For the \textit{neighbor prediction} task, the model is prompted to directly generate the texts of the neighbors of a given node based on its node encoding and text. This task is designed to further align the node encoding with the neighborhood of a node, and to enable the model to directly generate the neighbors of a node, which helps accelerate the inference process.

For the \textit{link prediction} task, the model is asked to predict whether two nodes are connected based on their texts, node encodings, and the pairwise encoding between them. Each instruction contains one source node and multiple candidate target nodes, with an equal number of positive and negative samples. This setup allows the model to compare the current candidate with other candidates in the context, enabling the model to have in-context learning capabilities during inference.

\mypar{Training schedule.} 
We adopt a two-stage tuning strategy. In stage 1, we focus on training the pairwise encodings and aligning the special encodings with the word embeddings of the pre-trained LLM. The LLM is kept frozen, and only the encoding-related modules, including the pairwise encoder and two alignment projectors, are tuned. In stage 2, we keep the encoding-related modules frozen and tune the LLM through LoRA \cite{Hu2021LoRALA} to improve its understanding of the encodings.

\subsection{Retrieval-Rerank Scheme for Inference}
\label{sec:rerank}

The inference stage of link prediction involves ranking a set of candidate target nodes $\mathcal{C}=\{t_1,t_2,\cdots, t_{N_C}\}$ for each source node. However, directly ranking all candidates using the LLM can be computationally expensive, especially when the candidate set $\mathcal{C}$ is large. To address this challenge, we propose a retrieval-rerank scheme that efficiently narrows down the candidate set before applying the LLM for final ranking.

\mypar{Retrieval stage.}
In the retrieval stage, we aim to quickly identify a smaller subset of $n_C\ll N_C$ candidate target nodes that are most likely to be connected to the source node. Inspired by GPT4Rec~\cite{li2023gpt4rec}
, we prompt the LLM to directly generate the texts of potential neighbors for the source node. The generated texts are then used as queries to search for relevant candidates in the target set $\mathcal{C}$ using the BM25
 algorithm~\cite{10.1561/1500000019}.

To ensure the generated texts cover a diverse range of content, we employ beam search with diversity penalty applied between each beam group~\cite{vijayakumar2018diverse}. Furthermore, we apply a distance-based grouping strategy to select $\beta n_C$ nodes from the source node's 2-hop neighborhood $\mathcal{N}_s^2$ and $(1-\beta)n_C$ nodes from the rest of the graph $\mathcal{V}\backslash\mathcal{N}_s^2$ with the highest BM25 scores. We set $\beta=0.65$ in our experiments.

\mypar{Rerank stage.}
In the rerank stage, we use the LLM to rank the retrieved $n_C$ candidates and obtain the final link prediction results. To quantify the probability of an edge existing between the source node $s$ and a candidate target node $t_i$, we introduce the Yes/No Index:
\begin{equation}
    \text{Yes/No Index} = \frac{p(\text{"Yes"}\mid \text{context})}{p(\text{"Yes"}\mid \text{context})+p(\text{"No"}\mid \text{context})}
\end{equation}

The probabilities are calculated by applying the softmax function to the logits output by the LLM.

To leverage the LLMs' in-context learning ability, we include several examples in the prompt's context, where positive samples are the source node's neighbors in the training set, and negative samples are randomly selected from $\mathcal{V}\backslash\mathcal{N}_s^1$. Since the prompts for all candidate target nodes share the same context about the source node and examples, we can save computation by reusing the keys and values of this shared part when calculating the logits for each candidate. The time complexity analysis is in Appendix \ref{sec:reuse_kv}.

The retrieval-rerank scheme significantly reduces the computational cost of link prediction inference while maintaining high accuracy, making \method more practical for real-world applications.

\section{Experiments}
\label{sec:experiments}
We conduct comprehensive experiments to evaluate the effectiveness and efficiency of our proposed framework and address the following research questions:

\begin{itemize}
    \item RQ1: Compared to other GAE-based and LLM-based baselines, what is the overall performance of our \method model in the link prediction task?
    \item RQ2: What is the zero-shot and few-shot generalization ability of our model?
    \item RQ3: How scalable and efficient is our model during inference, i.e., candidate target nodes ranking?
    \item RQ4: What is the individual contribution of the node and pairwise encodings to the overall performance of our model?
\end{itemize}

\subsection{Experimental Setup}

\mypar{Datasets.} Following Patton \cite{jin2023patton}, we train and evaluate our model on these four datasets: Amazon-Sports, Amazon-Clothing \cite{McAuley2015ImageBasedRO}, MAG-Geology, and MAG-Math \cite{10.1145/2740908.2742839, zhang2023effect}. The first two datasets are e-commerce networks, where each node represents a product on Amazon, and an edge between two nodes means that they are frequently purchased together. The last two datasets are academic networks, where each node represents a paper and an edge between two nodes means that one cites the other. Since the inference process of link prediction is computationally expensive for LLMs, we subsample all the datasets by randomly selecting 20,000 nodes. Detailed statistics about the datasets are reported in Appendix \ref{sec:statistics}.

\mypar{Tasks.} We leverage \method for two tasks: link prediction and neighbor prediction. The target of \textit{link prediction} is to rank the $N_C$ candidate target nodes based on their probabilities of being connected with the source node. We \textbf{do not} apply the retrieval stage in this experiment for a clearer assessment of the model's ranking capability. Besides, in the \textit{neighbor prediction} task, the model is prompted to directly generate the text of neighbors, which will then be used to retrieve $n_C$ candidate target nodes from a pool of $N_C$. Note that link prediction is the primary task, while neighbor prediction is formulated to make the link prediction more efficient during inference. 

\mypar{Metrics.} For the link prediction task, we use MRR and Hits@1, which are two of the most commonly used evaluation metrics in the literature~\cite{hu2020open, jin2023patton}. MRR is defined as the mean reciprocal rank of the positive target node, and Hits@1 is the proportion of times the positive target node is ranked top 1 among all candidates. For the neighbor prediction task, the generated texts are used as queries for retrieval, and the performance is evaluated using the MRR and runtime of the whole retrieval-rerank process. If the positive target node is not retrieved, its ranking is considered to be infinity.

\mypar{Baselines.} We apply three categories of competitive models. The first category is traditional GAE-based models, including GCN \cite{Kipf2016SemiSupervisedCW}, GraphSAGE \cite{Hamilton2017InductiveRL}, and GATv2 \cite{Brody2021HowAA}. The second category comprises models that make use of pairwise information, including SEAL \cite{Zhang2018LinkPB}, BUDDY \cite{chamberlain2023graph}, and LPFormer \cite{shomer2023adaptive}. The third category consists of LLM-based models like our \method, which includes vanilla LLaMA2-7B \cite{touvron2023llama}, GraphGPT \cite{tang2023graphgpt} and LLaGA \cite{chen2024llaga}. Note that only the vanilla LLaMA2-7B model is not trained on these datasets. More descriptions of baseline models are in Appendix \ref{sec:baselines}.

\mypar{Implementation details.} We leverage LLaMA2-7B \cite{touvron2023llama} as the backbone of our framework. In each stage of the instruction tuning, the link prediction and neighbor prediction tasks are carried out for 1 epoch respectively. Details about hyper-parameters and computing resources are in Appendix \ref{sec:implementation}. Furthermore, for a fair comparison, we employ bert-base-uncased~\cite{devlin2018bert} as the original embeddings of the nodes for the training of \method and all the GNN-based baseline models. For LLM-based models, we use the embedding methods described in their original papers.

\subsection{Overall Performance (RQ1)}

\label{sec:overall_performance}

\begin{table}[t!]
\caption{Performance of \method\ and baseline models on the link prediction task on 4 datasets. The best results are marked in \textbf{bold}.}
\centering
\resizebox{\textwidth}{!}
{
\begin{tabular}{l cc cc cc cc}
\toprule
Dataset        & \multicolumn{2}{c}{Amazon-Sports} & \multicolumn{2}{c}{Amazon-Clothing} & \multicolumn{2}{c}{MAG-Math} & \multicolumn{2}{c}{MAG-Geology} \\ 
\midrule
        & MRR              & Hits@1           & MRR               & Hits@1            & MRR               & Hits@1            & MRR            & Hits@1          \\ 
        \midrule
GCN~\cite{Kipf2016SemiSupervisedCW}            & 70.44   & 60.17     & 68.18   & 60.00     & 51.35       & 40.11      & 45.80     & 33.14             \\
GraphSAGE~\cite{Hamilton2017InductiveRL}      & 77.60    & 68.43     & 81.17   & 71.61     & 50.97       & 36.77      & 44.17     & 28.41          \\
GATv2~\cite{Brody2021HowAA}          & 81.44   & 72.96     & 87.83   & 81.56     & 65.65       & 55.27      & 51.59     & 37.95          \\

 \midrule

SEAL~\cite{Zhang2018LinkPB}           & 76.79   & 69.58   & 82.14     & 75.41     & 61.79       & 56.34      & 58.05     & 50.07     \\
BUDDY~\cite{chamberlain2023graph}          &  81.33  &  74.00  &  83.89  &  76.52   & 58.15   &  48.30     &   54.95   &  45.32      \\
LPFormer~\cite{shomer2023adaptive}       & 69.94   & 62.61   & 65.99     & 56.29     & 47.97       & 42.19      & 43.10     & 35.08    \\
\midrule
LLaMA2~\cite{touvron2023llama}         & 40.81   & 30.88   & 30.22     & 22.49     & 22.92       & 13.54      & 21.86     & 13.47     \\
GraphGPT~\cite{tang2023graphgpt}       & 14.82   &  5.96   &  32.29    & 14.28     &  12.43     &  4.37    &  9.78    &  2.62      \\
LLaGA~\cite{chen2024llaga}          & 83.41   & 75.40  & 84.49      & 77.54     & 74.25       & 63.34      & 62.19     & 49.80      \\ \midrule

\textbf{\method\ (Ours)} & \textbf{87.07} &\textbf{79.56} & \textbf{90.18} & \textbf{84.82} & \textbf{81.03}  & \textbf{71.01} & \textbf{75.43} & \textbf{64.57} \\ \bottomrule
\end{tabular}
}
\label{table:overall_perform}
\end{table}

First, we compare the performance of \method\ and state-of-the-art baselines on the link prediction task. The number of candidates $N_C$ is set to be 150 for all datasets.
The results are presented in Table \ref{table:overall_perform}.
The proposed \method\ model significantly outperforms all the baseline models across all four datasets on the link prediction task, achieving state-of-the-art performance in terms of both MRR and Hits@1 metrics. This highlights the effectiveness of \method's approach in leveraging node-wise, pairwise, and textual information for link prediction. 

While GraphGPT~\cite{tang2023graphgpt} performs relatively poorly as it is designed for node classification tasks, the vanilla LLaMA2~\cite{touvron2023llama}, though not specifically trained for this task, still demonstrates reasonable performance, suggesting the strong potential of pre-trained language models in utilizing textual attributes of TAGs. LLaGA~\cite{chen2024llaga}, which leverages both the node-wise structural and textual information, outperforms most baselines but still lags behind \method, indicating the importance of effectively modeling pairwise information for link prediction. Although GATv2~\cite{Brody2021HowAA} also achieves impressive performance thanks to its dynamic attention mechanism, overall, models that can capture pairwise information, especially BUDDY~\cite{chamberlain2023graph} and SEAL~\cite{Zhang2018LinkPB}, tend to perform better than traditional graph auto-encoder (GAE)-based ones.

\subsection{Generalization Ability (RQ2)}

\label{sec:generalization_ability}

\begin{table}[t!]
\caption{Zero-shot performance on the link prediction task. $A\rightarrow B$ indicates that the model is trained on dataset A and evaluated on an unseen dataset B. The best results are marked in \textbf{bold}.}
\centering
\resizebox{\textwidth}{!}
{
\begin{tabular}{l cc cc cc cc}
\toprule
& \multicolumn{4}{c}{Cross-category} &\multicolumn{4}{c}{Cross-domain}\\ 
Dataset        & \multicolumn{2}{c}{Sports$\rightarrow$Clothing} & \multicolumn{2}{c}{Math$\rightarrow$Geology} & \multicolumn{2}{c}{Sports$\rightarrow$Math} & \multicolumn{2}{c}{Math$\rightarrow$Sports} \\ 
\midrule
        & MRR              & Hits@1           & MRR               & Hits@1            & MRR               & Hits@1            & MRR            & Hits@1          \\ 
        \midrule
GraphSAGE~\cite{Hamilton2017InductiveRL}      &  51.88  & 36.43   & 8.11   &  2.06    &   19.27     &     8.88     &   27.64     &    14.52            \\
GATv2~\cite{Brody2021HowAA}      &  67.01 & 55.36     &   21.41  &  13.54    &   36.87         &   26.86         &   49.68     &    37.21          \\
BUDDY~\cite{chamberlain2023graph}          & 81.23   & 72.01   & 24.45   &  13.35    &      26.43  &    15.73     &  42.39    &  21.22        \\
LPFormer~\cite{shomer2023adaptive}          &  68.26  &  59.64  &   42.70  &  35.08   &      48.50 &   43.75    &   56.78    &  44.13       \\
LLaGA~\cite{chen2024llaga}          &  78.62  & 68.90    & 23.20     & 13.13     &  23.86     & 12.97     &   53.67 &   40.46   \\ \midrule

\textbf{\method\ (Ours)} & \textbf{85.74} &\textbf{79.01} & \textbf{71.58} & \textbf{60.42} & \textbf{67.50}  & \textbf{55.27} & \textbf{69.38} &   \textbf{55.40} \\ \bottomrule
\end{tabular}
}
\label{table:generalization}
\end{table}

We then evaluate the generalization ability of \method in two settings: zero-shot generalization and few-shot in-context learning.

\mypar{Zero-shot generalization.}
We train \method on one dataset $A$ and evaluate it on another dataset $B$, without fine-tuning or additional examples in the prompt.
Results are summarized in Table~\ref{table:generalization}.
Sports$\rightarrow$Clothing and Math$\rightarrow$Geology are cross-category generalization, since both datasets are from the same domain, such as e-commerce networks or academic networks.
Sports$\rightarrow$Geology and Math$\rightarrow$Clothing are cross-domain generalization, where the two datasets are from different domains.

The proposed \method model demonstrates the strongest generalization ability, outperforming all other baseline models in all settings. \method not only benefits from its LLM backbone, which possesses powerful text feature generalization capabilities, but also from effectively capturing the structural commonalities of link formation through the node and pairwise encodings. Models that make use of pairwise information in the table all demonstrate impressive generalization abilities, but they lag behind \method due to their inability to transfer knowledge of text features. Additionally, although LLaGA~\cite{chen2024llaga} also possesses LLM's text generalization capability, it underperforms since it fails to capture the pairwise structural information explicitly and effectively.

\begin{figure}[t]
    \centering
    \hfill
    \begin{subfigure}[b]{0.4\textwidth}
        \centering
        \includegraphics[width=\textwidth]{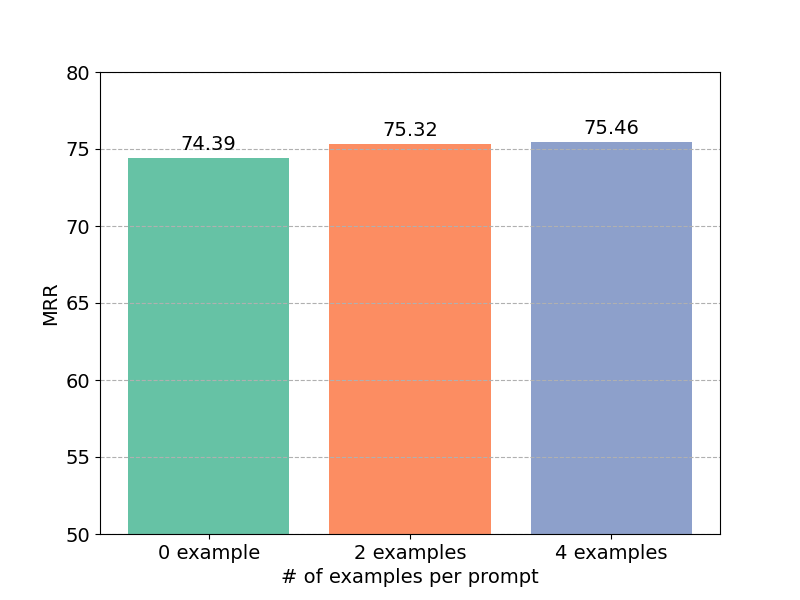}
        \caption{MAG-Geology}
    \end{subfigure}
    \hfill
    \begin{subfigure}[b]{0.4\textwidth}
        \centering
        \includegraphics[width=\textwidth]{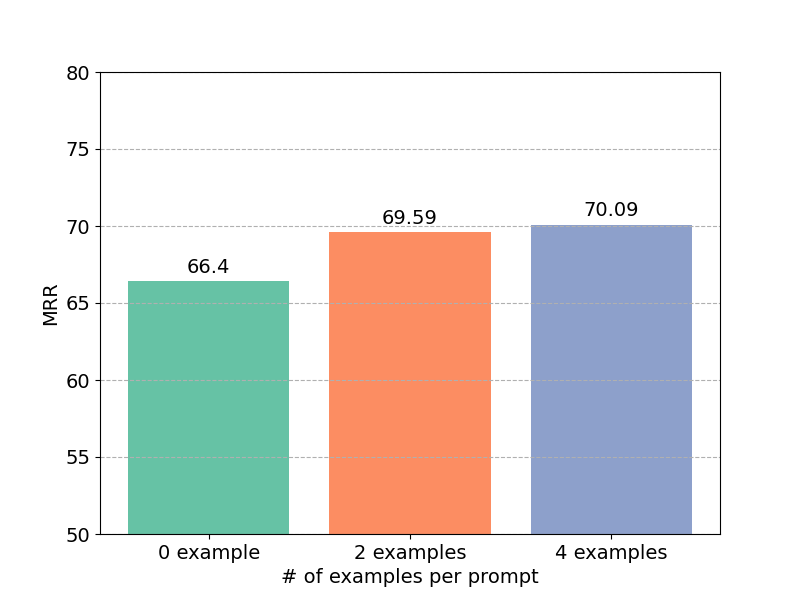}
        \caption{Amazon-Sports $\rightarrow$ MAG-Geology}
    \end{subfigure}
    \hfill
    \caption{Performance of few-shot in-context learning. \textbf{(Left):} When the model is fine-tuned and evaluated on MAE-Geology, in-context learning does not significantly improve the performance. \textbf{(Right):} For cross-domain generalization, in-context learning boosts the performance significantly.}
    \label{fig:num_examples}
\end{figure}

\mypar{Few-shot in-context learning.}
We also evaluate the generalization ability of \method\ with the help of a few examples in test time. Other baseline models do not have the capability for in-context learning due to their architectures or prompt designs. The results are shown in Figure~\ref{fig:num_examples}.
When we train and evaluate on the same dataset, in-context learning is not very helpful since the model has been fine-tuned on the dataset.
However, for cross-domain generalization, in-context learning is extremely helpful and boosts the performance from 66.40 to 70.09 on the MAG-Geology dataset.

\subsection{Scalability Analysis (RQ3)} 

\label{sec:scalability}

In this section, we investigate the scalability of \method by evaluating the performance and runtime of \method and other LLM-based model with large candidate target node set $\mathcal{C}$.

\begin{figure}[t]
    \centering
    \subfloat[Amazon-Sports]{\includegraphics[width=0.25\linewidth]{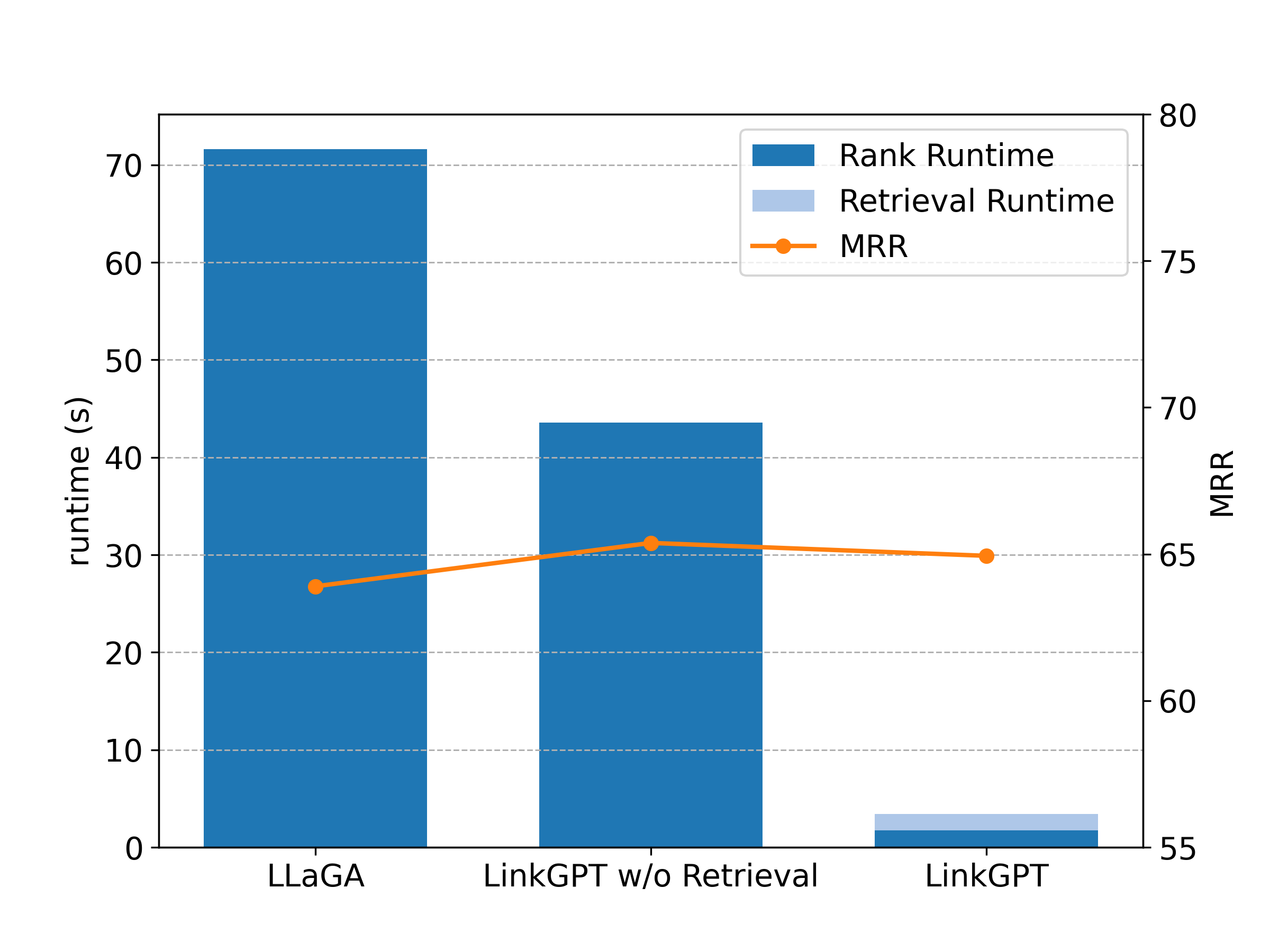}}
    \subfloat[Amazon-Clothing]{\includegraphics[width=0.25\linewidth]{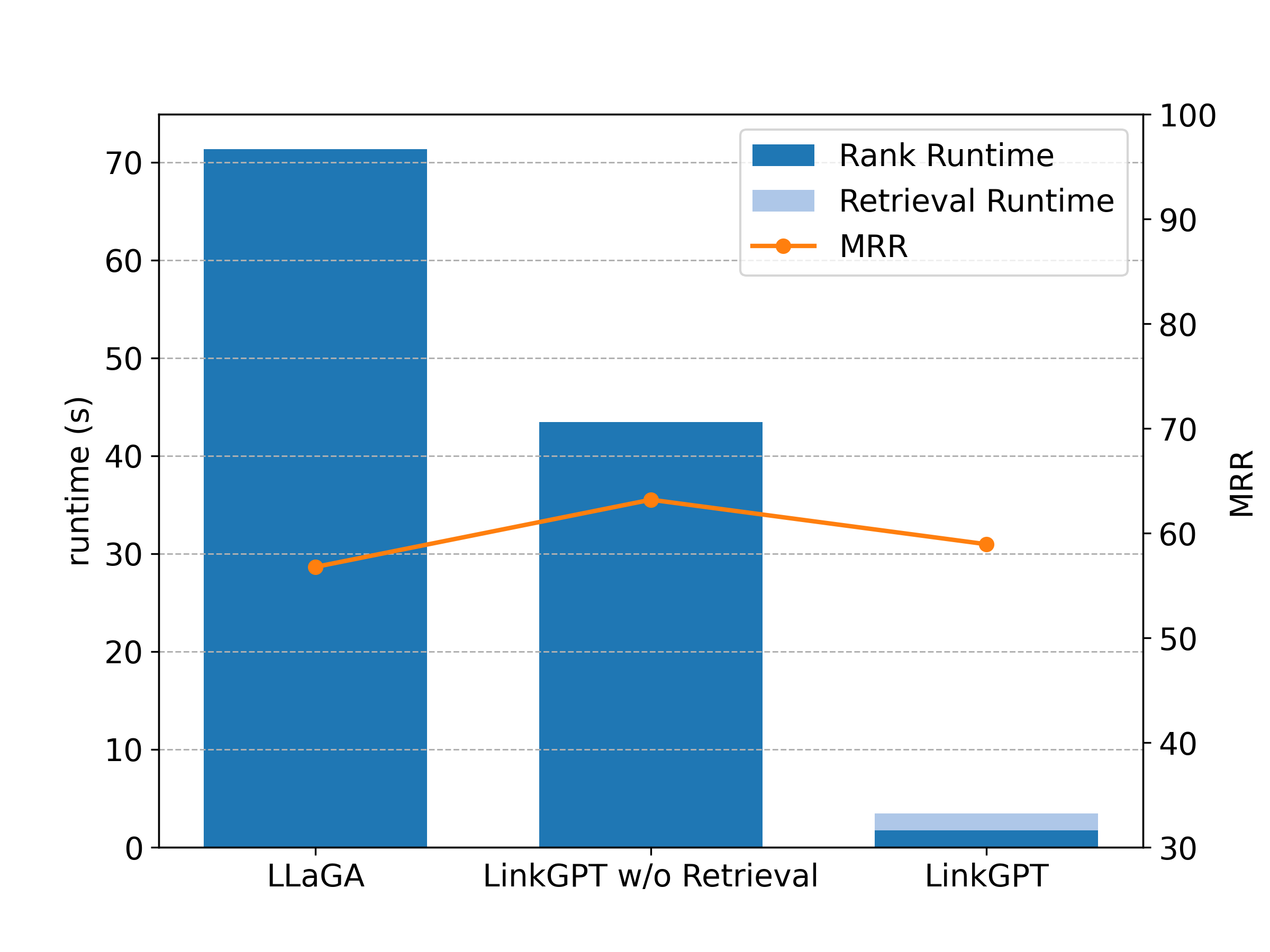}}
    \subfloat[MAG-Math]{\includegraphics[width=0.25\linewidth]{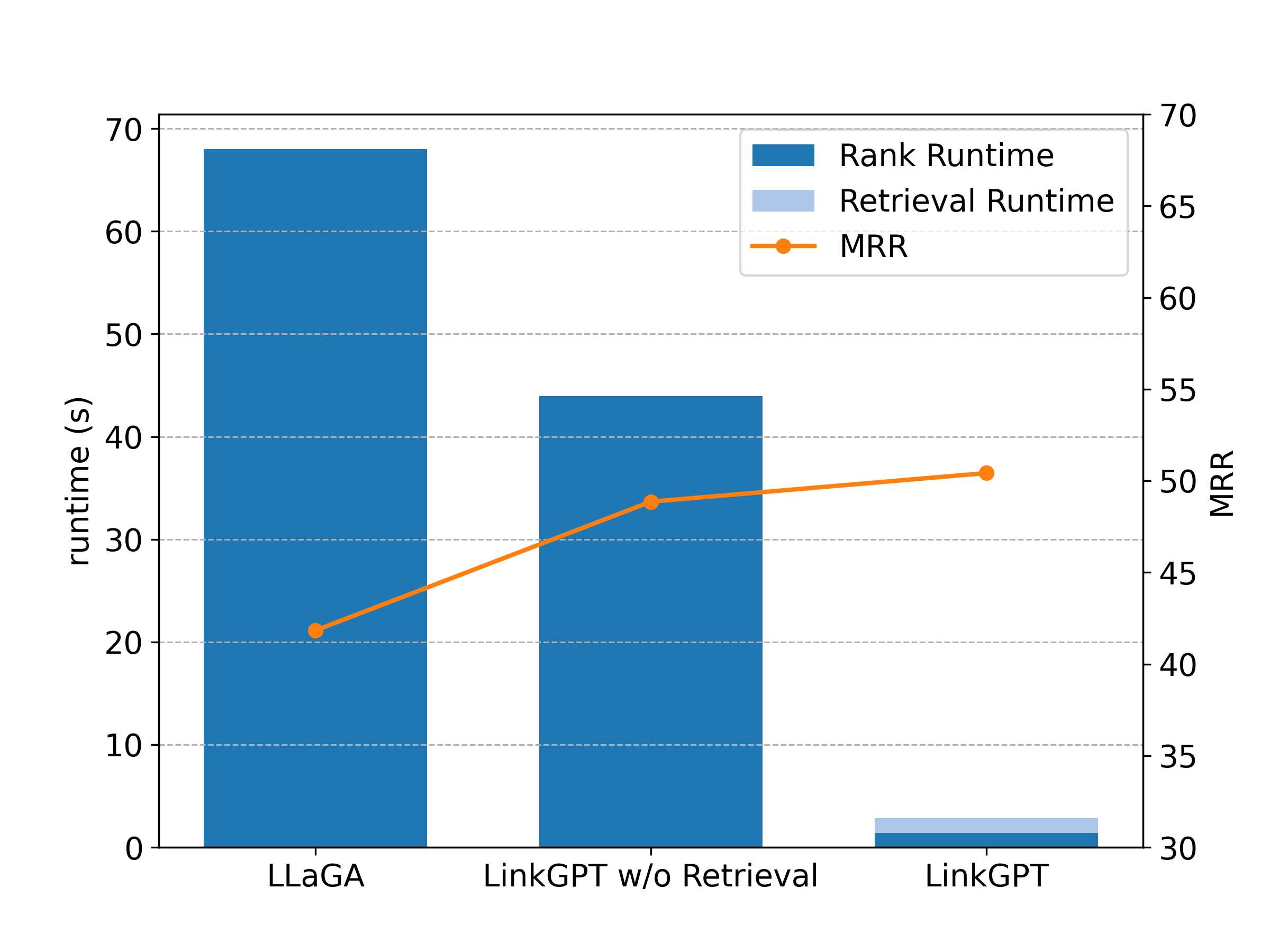}}
    \subfloat[MAG-Geology]{\includegraphics[width=0.25\linewidth]{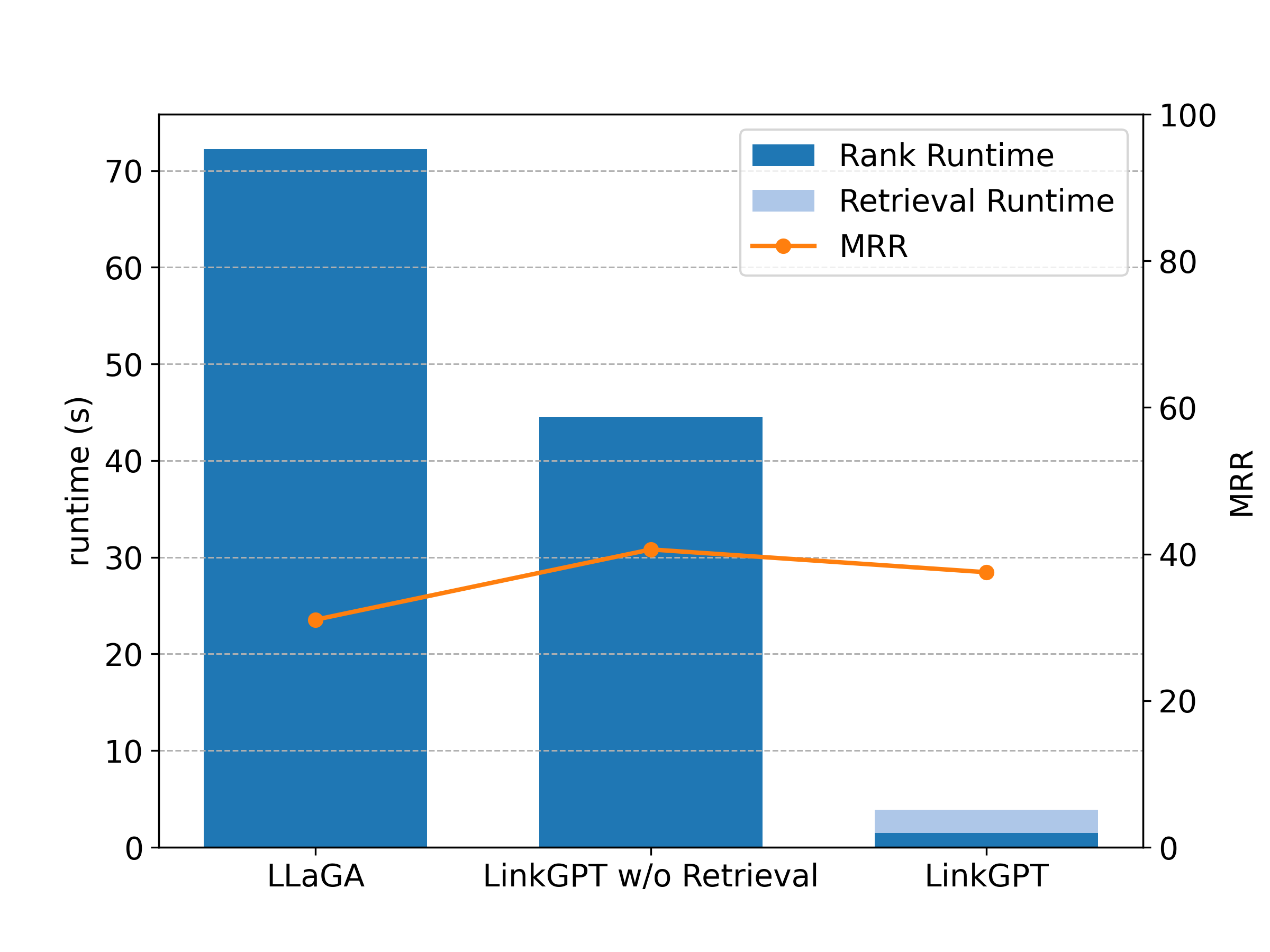}}
    \caption{Performance of the retrieval-rerank scheme of \method. The line chart represents the final MRR, using the right axis, while the bar chart represents the total time required for each $(s, \mathcal{C})$ pair during inference, using the left axis. Here $s$ denotes the source node and $\mathcal{C}$ denotes the candidate target node set.}
    \label{fig:runtime_mrr}
\end{figure}

\mypar{Setup.} We set the number of negative candidates $N_C=1,800$ for all four datasets, the number of candidates to retrieve $n_C=30$ for Amazon, and $n_C=60$ for MAG. Due to the computational costs, we randomly selected 200 pairs of source node $s$ and candidate target node set $\mathcal{C}$ for the experiments. Furthermore, we conducted experiments under three settings: LLaGA~\cite{chen2024llaga}, \method w/o retrieval, and \method. The inference of \method consists of two stages: retrieval and rerank. For the neighbor prediction in the retrieval stage, we set 5 beam groups, with each group having a size of 3.

\mypar{Results.}
As shown in Figure \ref{fig:runtime_mrr}, \method w/o Retrieval significantly outperforms LLaGA~\cite{chen2024llaga} in terms of MRR across all datasets and has relatively lower runtime. Thanks to \method's neighbor prediction capability, the inclusion of the retrieval stage can improve the inference efficiency by 10 times while ensuring that the MRR decline does not exceed 5\%. Furthermore, \method's MRR is still not lower than LLaGA's across all datasets, showing the effectiveness and efficiency of the retrieval-rerank scheme of \method. Case study of the neighbor prediction task is in Appendix \ref{sec:case_study}.

\begin{table}[h]
\centering
\caption{Ablation on the node encoding and pairwise encoding. The pairwise encoding are more important than node encoding for the link prediction task, since pairwise information is more helpful. However, both encodings contribute to the final performance of \method.}
\vspace{0.5em}
\begin{tabular}{l cc cc}
\toprule
Dataset        & \multicolumn{2}{c}{Amazon-Clothing} & \multicolumn{2}{c}{MAG-Geology} \\ \midrule
               & MRR     & Hits@1     & MRR     & Hits@1       \\ \midrule
\textbf{\method\ (Ours)} & \textbf{90.18} &\textbf{84.82} & \textbf{75.43}  & \textbf{64.57} \\ 
- w/o node encoding       & 88.77   & 82.46     & 74.73   &  63.50      \\
- w/o pairwise encoding   & 83.08   & 74.11     & 72.07   &  60.42      \\
\bottomrule
\end{tabular}
\label{table:ablation}
\end{table}

\subsection{Ablation Study (RQ4)}

\mypar{Setup.}
In this section, we investigate the individual contributions of the node and pairwise encodings. We remove each encoding component separately and train the model using the same two-stage instruction tuning strategy. This allows us to isolate the effect of each encoding on the model's performance.
For the ablated models, we maintain the same hyper-parameters and training settings as the full \method model to ensure a fair comparison. The models are evaluated on the same benchmark datasets, and their link prediction performance is reported.

\mypar{Results.}
The results of the ablation study are presented in Table \ref{table:ablation}. Incorporating the node and pairwise encodings leads to significant improvements in link prediction performance. This highlights the importance of explicitly encoding node and pairwise information to enhance the LLMs' understanding of the graph structure. The node encoding allows the model to capture node-wise structural information, such as the features of its neighborhood, and the pairwise encoding enables the model to reason about the relationships between node pairs.

Comparing the contributions of the two encodings, we find that the pairwise encoding plays a more crucial role in the link prediction task. The removal of pairwise encoding results in a larger performance drop compared to removing the node encoding. This observation aligns with the intuition that pairwise information, which captures the connectivity patterns between nodes, is more directly relevant to predicting missing links.

\subsection{Limitations}

\label{sec:limitations}

Finally, although \method brings LLM to the link prediction and achieves state-of-the-art performance, it also brings certain limitations.

\mypar{The balance of performance and efficiency.}
One of the primary bottlenecks of adapting LLM to link prediction is efficiency, as link prediction typically has hundreds of negative targets to rank.
We have proposed a retrieval-rank scheme to address it.
However, compared with LLaGA~\cite{chen2024llaga},
\method\ experiences a more noticeable performance decline when the target candidate set $\mathcal{C}$ constitutes an excessively large proportion of the entire node set $\mathcal{V}$.
On MAG-Math, when $N_C/|\mathcal{V}|$ increases from 5\% to 15\%, the MRR of \method w/o retrieval drops from 58.95 to 38.82, while that of LLaGA~\cite{chen2024llaga} only drops from 47.83 to 32.90.
We believe when $N_C/|\mathcal{V}|$ is too large, many nodes that are close to the source node $s$ are selected as negative target candidates. These nodes have a high probability of being connected to $s$ due to pairwise encoding, leading to lower rankings of positive candidates.

\mypar{LLM hallucinations.}
One key limitation is the susceptibility of LLM to hallucinations, where the model generates plausible but incorrect information. LLMs are known to sometimes produce outputs that are inconsistent with the input data or introduce false statements presented as facts~\cite{ji2023survey}. This can be particularly problematic in the context of link prediction, as hallucinated links could lead to erroneous conclusions and decision-making based on the graph data.
To mitigate the impact of hallucinations, future work could explore techniques such as retrieval-augmented generation~\cite{tonmoy2024comprehensive}, where the LLM is provided with relevant external knowledge to ground its outputs in factual information. Incorporating explicit knowledge bases or structured data alongside the LLM could help reduce the occurrence of hallucinations. Additionally, implementing output filtering mechanisms or confidence thresholds to identify and discard potentially hallucinated links could improve the reliability of the predictions. 
\section{Conclusion}
\label{sec:conclusion}
In this paper, we introduced \method, a novel approach that leverages the power of Large Language Models (LLMs) for link prediction tasks on text-attributed graphs (TAGs). By combining node and pairwise encodings, two-stage fine-tuning, and a retrieval-rerank scheme for inference, \method effectively addresses the challenges of incorporating pairwise node information and efficiently ranking a large number of candidates. Our extensive experiments on real-world datasets demonstrate \method's superior performance and computational efficiency compared to state-of-the-art methods. Furthermore, the cross-category generalization experiments highlight \method's robustness and ability to transfer knowledge across different categories and domains.

\mypar{Social impacts.}
\label{sec:social}
\method has the potential to revolutionize various real-world applications that rely on graph-structured data. For instance, in social networks, \method can help identify potential connections between users, fostering community growth and enhancing user experience. In recommendation systems, \method can improve the accuracy of personalized recommendations by discovering hidden preferences and interests. Moreover, in biological networks and knowledge graphs, \method can aid in uncovering novel relationships between entities, leading to groundbreaking discoveries in fields such as drug discovery and scientific research. However, it is crucial to consider the ethical implications of using LLMs for link prediction, such as privacy concerns and the potential for biased predictions. Future research should focus on developing techniques to mitigate these risks and ensure the responsible deployment of \method in real-world scenarios.

\balance
%

\bibliographystyle{plain}

\clearpage
\appendix

\label{sec:appendix}
\section{Details of the Neighborhood-Aware Node Encoder}
\label{sec:graphformers}
We employ GraphFormers \cite{Yang2021GraphFormersGT} as the node encoder. GraphFormers is a transformer encoder with a GNN applied between each block, and the text encoding and graph aggregation are thus alternately performed. Formally, the forward pass of each layer in GraphFormers is given by
\begin{equation}
z_v^{(l)}=\text{GNN}(\{\mathbf{H}_u^{(l)}[\text{CLS}]\mid u\in\mathcal{N}_v^1\}) 
\end{equation}

\begin{equation}
    \tilde{\mathbf{H}}_v^{(l)}=\text{concat}(z_v^{(l)},\mathbf{H}_v^{(l)})
\end{equation}

\begin{equation}
\tilde{\mathbf{H}}_v^{(l)'}=\text{LayerNorm}(\mathbf{H}_v^{(l)}+\text{MHA}_{\text{asy}}(\tilde{\mathbf{H}}_v^{(l)}))
\end{equation}

\begin{equation}
\mathbf{H}_v^{(l+1)}=\text{LayerNorm}(\tilde{\mathbf{H}}_v^{(l)'}+\text{MLP}(\tilde{\mathbf{H}}_v^{(l)'}))
\end{equation}

where $\mathbf{H}_v^{(l)}$ is the hidden states in the $l$ -th layer of node $v$, and $\text{MHA}_{\text{asy}}$ means asymmetric multihead attention. The initial embedding $\mathbf{H}_v^{(0)}$ is obtained through a learnable embedding module. 

\begin{figure}[h]
    \centering
    \begin{subfigure}[b]{0.45\textwidth}
        \centering
        \includegraphics[width=\textwidth]{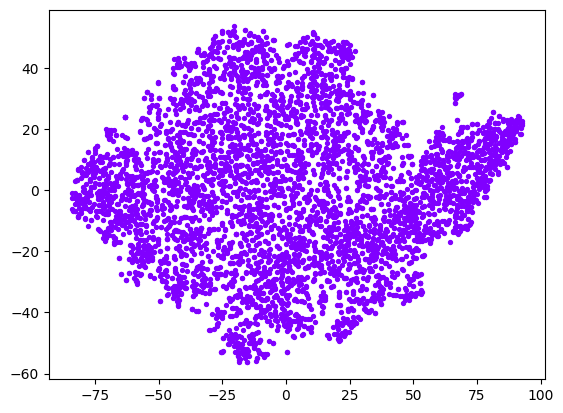}
        \caption{BERT}
    \end{subfigure}
    \hfill
    \begin{subfigure}[b]{0.45\textwidth}
        \centering
        \includegraphics[width=\textwidth]{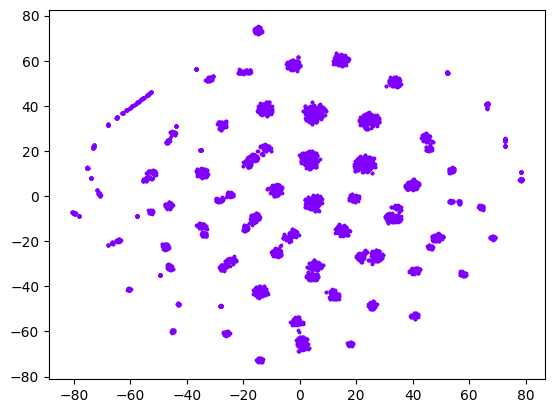}
        \caption{CGTP}
    \end{subfigure}
    \caption{Visualization of two node encoding methods.}
    \label{fig:vis_emb}
\end{figure}

We use the Amazon Sports dataset as an example to investigate the effect of contrastive graph-text pretraining (CGTP). We reduce the dimensions of node encoding to two dimensions for visualization through t-SNE \cite{Maaten2008VisualizingDU}, as shown in Figure~\ref{fig:vis_emb}. After CGTP, connected nodes tend to be closer in the embedding space, forming clear clusters. In contrast, embeddings produced by BERT are generally evenly dispersed throughout the space, showing a weaker tendency to form clusters. This indicates that our node encoding can effectively understand the central node by leveraging information from its neighborhood.

\section{Details of Pairwise Encoding}
\label{sec:pairwise}
Pairwise information plays a crucial role in the task of link prediction \cite{Mao2023RevisitingLP}. LPFormer is an adaptive pairwise encoder, and many common pairwise heuristics, such as Common Neighbors \cite{Newman2001ClusteringAP} and Katz Index \cite{Katz1953ANS}, can be represented as its special cases. Specifically, the pairwise encoding $f_P(a, b)$ between nodes $a$ and $b$ is calculated as follows:

\begin{equation}
    f_P(a,b)=\sum_{u\in\mathcal{V}_{(a,b)}}w(a,b,u)\odot h(a,b,u)
\end{equation}

where $w(a,b,u)$ and $h(a,b,u)$ denote the importance and encoding of node $u$ relative to the relation between $a$ and $b$ respectively, and $\mathcal{V}_{(a,b)}$ denotes the set of notes that might be important to the relation between $a$ and $b$. Specifically, $\mathcal{V}_{(a,b)}$ consists of nodes with Personalized PageRank (PPR) \cite{Brin1998TheAO} score higher than a threshold $\eta$ relative to either $a$ or $b$.

\begin{equation}
    \mathcal{V}_{(a,b)}=\{u\in\mathcal{V}\mid \text{ppr}(a,u)\geq \eta, \text{ppr}(b,u)\geq \eta\}
\end{equation}

In this paper, we set $\eta=0.01$ for nodes in $\mathcal{N}_a^1 \cup \mathcal{N}_b^1$, and $\eta=1$ for all other nodes to filter them out. Moreover, $w(a,b,u)$ and $h(a,b,u)$ are given by:

\begin{equation}
    \tilde{w}(a,b,u)=\phi(\mathbf{h}_a, \mathbf{h}_b, \mathbf{h}_u, \mathbf{rpe}_{(a,b,u)})
\end{equation}

\begin{equation}
    w(a,b,u)=\frac{\exp(\tilde{w}(a,b,u))}{\sum_{v\in\mathcal{V}\backslash\{a,b\}}\exp(\tilde{w}(a,b,u))}
\end{equation}

\begin{equation}
    h(a,b,u)=\mathbf{W}\text{concat}(\mathbf{h}_u, \mathbf{rpe}_{(a,b,u)})
\end{equation}

where $\phi$ denotes the GATv2 attention mechanism \cite{Brody2021HowAA}, $\mathbf{W}$ is a trainable weight matrix, and $\mathbf{rpe}_{(a,b,u)}$ denotes the relative positional encoding (RPE), which is given by $\mathbf{rpe}_{(a,b,u)}=\text{MLP}(\text{ppr}(a,u),\text{ppr}(b,u))$. The learned pairwise encoding $s(a,b)$ is then mapped to the semantic space of the LLM using a simple alignment projector $\sigma_{\text{P}}$, which can be either a linear layer or a small MLP.

\section{Time Complexity Analysis of Reusing Keys and Values}

\label{sec:reuse_kv}

Denote the length of the part about the source node and all examples by $m_s$, and denote the length of the part about the candidate target node by $m_t$. Since the time complexity of each self-attention layer in LLM is $O(m^2)$, where $m$ represents the length of the input sequence, if we do not reuse any past keys or values, the time complexity for all $n_C$ instructions would be

\begin{equation}
    O(n_C(m_s+m_t)^2)=O(n_C m_s^2 + n_C m_s m_t + n_C m_t^2)
\end{equation}

However, since for each $(s, \mathcal{C})$ pair, where $s$ denotes the source node and $\mathcal{C}$ denotes the candidate set, all instructions share the same part about the source node and examples. By reusing the keys and values generated by each self-attention layer of the shared part, we can reduce the time complexity to

\begin{equation}
    O\left(m_s^2 + n_C\sum_{i=1}^{m_t}(m_s+i)\right)=O(m_s^2+n_C m_s m_t + n_C m_t^2)
\end{equation}

Due to the existence of examples and question text, $m_s$ is usually much longer than $m_t$. Therefore, reusing keys and values can boost the efficiency of the ranking stage of \method without any performance loss.

\section{Statistics of the Datasets}
\label{sec:statistics}
Since the training and inference of LLMs are both computationally expensive, we randomly subsample all datasets, retaining about 20k nodes and their interconnecting edges. The statistics of the subsampled datasets are shown in Table \ref{table:dataset_stats}. Furthermore, the data split ratio for training, validation, and test is 9:0.5:0.5 due to the high computation cost of inference.

\begin{table}[]
\centering
\caption{Statistics of the four datasets.}
\begin{tabular}{lcccc}
\toprule
            & Amazon-Sports & Amazon-Clothing & MAG-Mathematics & MAG-Geology \\ \midrule
\# of Nodes & 20,417        & 20,180          & 19,878          & 20,530      \\
\# of Edges & 48,486        & 44,775          & 34,676          & 51,540        \\ \bottomrule
\end{tabular}
\label{table:dataset_stats}
\end{table}

\section{Descriptions of the Baseline Models}
\label{sec:baselines}

\subsection{Traditional GAE-Based Models}

 For these models, we use the implementation of the Deep Graph Library (DGL) \cite{wang2020deep}. After obtaining the embedding of each node through the model, a 3-layer MLP is applied to determine whether two nodes are connected.

\mypar{GCN}~\cite{Kipf2016SemiSupervisedCW} GCN extends CNN to graph-structured data, learning node features through information passing and aggregation between nodes and their neighbors.

\mypar{GraphSAGE}~\cite{Hamilton2017InductiveRL} GraphSAGE learns graph representations by sampling and aggregating features from a node's neighbors, which makes it particularly suitable for large graphs.

\mypar{GATv2}~\cite{Brody2021HowAA} GATv2 uses a dynamic attention mechanism to assign different weights to each node in the graph, thus capturing important relationships between nodes.

\subsection{Models Leveraging Pairwise Information}

These models learn and utilize pairwise information between nodes in various ways. We use their respective official code for implementation.

\mypar{SEAL}~\cite{Zhang2018LinkPB} SEAL encodes local subgraphs between node pairs using GNNs, thus capturing richer structural information.

\mypar{BUDDY}~\cite{chamberlain2023graph} BUDDY uses subgraph sketches as message passing without explicit subgraph construction and applies feature precomputation technique, making it scalable in both time and space.

\mypar{LPFormer}~\cite{shomer2023adaptive} LPFormer designs an adaptive pairwise encoding using the attention mechanism of GATv2, thus effectively capturing pairwise information between nodes. 

\subsection{LLM-Based Models}

For LLaMA2, we use the transformers library released by Hugging Face~\cite{wolf2020huggingfaces} for implementation. For LLaGA and GraphGPT, we use their official code for implementation.

\mypar{LLaMA2}~\cite{touvron2023llama} The LLaMA2 model is a large autoregressive language model trained on a large corpus. We use the LLaMA2-7B version.

\mypar{GraphGPT}~\cite{tang2023graphgpt} GraphGPT performs text-graph grounding through contrastive learning and uses graph matching, node classification, and link prediction as training tasks for dual-stage instruction tuning. Moreover, it is primarily designed for node classification.

\mypar{LLaGA}~\cite{chen2024llaga} LLaGA enables LLMs to understand graph structural information by reorganizing nodes into two kinds of structure-aware sequence. It is trained using three tasks: node description, node classification, and link prediction.

\section{Implementation Details of \method}

\label{sec:implementation}

\begin{table}[]
\centering
\caption{Summary of parameters, memory usage, and training time for each stage. For \# of params tuned in stage 1, number 5.26 M is for linear projector, and 5.85 M is for 2-layer MLP as the projector.}
\begin{tabular}{lcccc}
\toprule
            & \# of Total Params& \# of Params Tuned & Memory & Time \\ \midrule
Stage 1 & 6.7 B & 5.26 M or 5.85 M        & 27 GB          & 2.2 hours              \\
Stage 2 & 6.7 B & 8.39 M        & 27 GB          & 2.8 hours                \\ \bottomrule
\end{tabular}
\label{table:computing}
\end{table}

In stage 1 of instruction tuning, only the pairwise encoder $f_P$, projector $\sigma_{\text{P}}$ and the node projector $\sigma_{\text{G}}$ are tuned. In stage 2, the aforementioned modules are frozen and only the LLM is tuned. When tuning the LLM, we apply LoRA \cite{Hu2021LoRALA} to the query and value projection modules with rank $r=16$ and alpha $\alpha=32$. During the whole training process, the learning rate is set to 3e-4 and the batch size is set to 8. Furthermore, we use AdamW~\cite{loshchilov2019decoupled} as the optimizer.

One NVIDIA A40 GPU is used for training and inference. The numbers of tuned parameters, memory usage, and training time for each stage of the instruction tuning are shown in Table \ref{table:computing}. The time of inference needed to produce the results of \method in Table \ref{table:overall_perform}, Table \ref{table:generalization} and Figure \ref{fig:num_examples} is approximately 2.3 hours for each dataset. The time of inference needed to produce the results of \method in Figure \ref{fig:runtime_mrr} is approximately 2.75 hours for each dataset.

\section{Licenses of Assets Used}

\label{sec:licences}

\mypar{LLaMA2}~\cite{touvron2023llama} LLaMA2 is a series of Large Language Models released by Meta. The LLaMA2-7B version is used in this paper. It is accessed through its \href{https://huggingface.co/meta-llama/Llama-2-7b-hf}{homepage} on Hugging Face, and its license is Llama 2 Community License Agreement. 

\mypar{MAG}~\cite{10.1145/2740908.2742839, zhang2023effect} The MAG dataset is an academic network released by Sinha et al. and organized by Zhang et al. It is accessed through its \href{https://github.com/yuzhimanhua/MAPLE}{official repository}, and is licensed under ODC-By.

\mypar{Amazon Products}~\cite{McAuley2015ImageBasedRO} The Amazon Products dataset is an e-commerce network released by Julian et al. It is accessed through its \href{https://jmcauley.ucsd.edu/data/amazon/links.html}{homepage}, and its license is MIT. The 2013 version is used in this paper. 

\section{Case Study of Neighbor Prediction Task}

\label{sec:case_study}

In this section, we discuss some cases of the neighbor prediction task. In the neighbor prediction task, the models are asked to directly generate the neighbors of a given node without any examples in the prompt. The generated text will then be used as the queries for retrieval.

Figure \ref{fig:geology_case} showcases an example from MAG-Geology~\cite{10.1145/2740908.2742839, zhang2023effect}, where the five sentences in the Generated Neighbors part represent the top-scoring sentences from the five beam groups (typically, there is little difference between sentences within the same beam group). In the figure, words that appear in both the positive target node and the generated texts are highlighted using the same colors. We can see that although there is no perfect matching, \method successfully predicts two important keywords: "magma" and "eruption." These two words did not appear in the source node or its neighbors but were mentioned multiple times in the generated texts, demonstrating the effectiveness of \method in the neighbor prediction task. 

Figure \ref{fig:sports_case} showcases an example of the Amazon-Sports~\cite{McAuley2015ImageBasedRO} dataset, where \method precisely outputs the text of the positive target node. We believe the reason is that, compared to the titles of academic papers, the titles of products on Amazon are usually simpler and shorter, making them easier to predict. Meanwhile, we also observe that some texts generated by the model do not actually exist in the dataset, or they differ from similar texts in the dataset to some extent. Such phenomenon indicates that the hallucination problem of LLMs also exists in \method. However, by introducing a frequency-based retriever, e.g., BM25~\cite{10.1561/1500000019}, the negative impact of hallucinations can be alleviated.

\begin{figure}[t]
\centering
\begin{mybox}
\textbf{Source Node:}

\vspace{0.25em}

An experimental study of the kinetics of decompression-induced crystallization in silicic melt.

\vspace{0.5em}

\textbf{Positive Target Node:}

\vspace{0.25em}

Preeruption conditions and timing of dacite‐andesite \textcolor{deepgreen}{\textit{magma}} mixing in the 2.2 ka \textcolor{deepred}{\textit{eruption}} at Mount Rainier.

\vspace{0.5em}

\textbf{Generated Neighbors:}

\vspace{0.25em}

Experimental study of the kinetics of decompression-induced crystallization in silicic \textcolor{deepgreen}{\textit{magma}}.

\vspace{0.25em}

Thermodynamics of decompression-induced crystallization in silicic \textcolor{deepgreen}{\textit{magma}}.

\vspace{0.25em}

Hydrothermal alteration of the 1980 Mt. St. Helens \textcolor{deepred}{\textit{eruption}}.

\vspace{0.25em}

Crystallization kinetics of silicic melt at 1200–1400 °C and 10–15 kbar: Implications for \textcolor{deepgreen}{\textit{magma}} chamber dynamics.

\vspace{0.25em}

Decompression-induced crystallization in silicic melt.

\end{mybox}
\caption{One example of the neighbor prediction task on the MAG-Geology dataset. The words that appear both in the positive target node and the generated neighbors are marked with the same colors.}
\label{fig:geology_case}
\end{figure}

\begin{figure}[t]
\centering
\begin{mybox}
\textbf{Source Node:}

\vspace{0.25em}

Louis Garneau Mistral Vent Jersey - Sleeveless.

\vspace{0.5em}

\textbf{Positive Target Node:}

\vspace{0.25em}

\textcolor{deepgreen}{\textit{K-Edge Handlebar Computer Mount for Garmin - }}\textcolor{deepred}{\textit{Men's}}.

\vspace{0.5em}

\textbf{Generated Neighbors:}

\vspace{0.25em}

\textcolor{deepgreen}{\textit{K-Edge Handlebar Computer Mount for Garmin - Men's}}.

\vspace{0.25em}

Pearl Izumi \textcolor{deepred}{\textit{Men's}} Elite Gel Glove.

\vspace{0.25em}

Louis Garneau \textcolor{deepred}{\textit{Men's}} 0 Calories Cycling Jersey.

\vspace{0.25em}

Lizard Skins DSP 2.5mm Bar Tape.

\vspace{0.25em}

BDI \textcolor{deepred}{\textit{Men's}} Classic Cycling Jersey. Made in Italy.

\end{mybox}
\caption{One example of the neighbor prediction task on the Amazon-Sports dataset. The words that appear both in the positive target node and the generated neighbors are marked with the same colors.}
\label{fig:sports_case}
\end{figure}

\section{Experiments on Other Datasets}

\begin{table}[t!]
\caption{Performance of \method\ and baseline models on the link prediction task on PubMed and ogbn-Arxiv. The best results are marked in \textbf{bold}.}
\centering

\begin{tabular}{l cc cc}
\toprule
Dataset        & \multicolumn{2}{c}{PubMed} & \multicolumn{2}{c}{ogbn-Arxiv} \\ 
\midrule
        & MRR              & Hits@1           & MRR               & Hits@1             \\ 
        \midrule
GATv2~\cite{Brody2021HowAA}     & 39.33   & 26.19 & 63.85 & 52.09  \\
LPFormer~\cite{shomer2023adaptive}       &  30.08  & 20.78  &   58.51   & 50.55  \\
LLaGA~\cite{chen2024llaga}          & 43.44 & 29.05  &  70.78    &  59.80 \\ \midrule

\textbf{\method\ (Ours)} & \textbf{66.54} &\textbf{53.42} & \textbf{81.03} & \textbf{72.06} \\ \bottomrule
\end{tabular}

\label{table:new_ds}
\end{table}

Since LLaGA~\cite{chen2024llaga} in not applied to the Amazon~\cite{McAuley2015ImageBasedRO} and MAG~\cite{10.1145/2740908.2742839, zhang2023effect} datasets in the original paper, we train \method and other baseline models on the PubMed~\cite{yang2016revisiting} and ogbn-Arxiv~\cite{hu2021open} datasets for a fairer and more comprehensive comparison. We use the publicly available checkpoint released by the authors of LLaGA. Due to the size of the ogbn-Arxiv dataset, 20,000 nodes are randomly selected for this experiment. As shown in Table \ref{table:new_ds}, \method outperforms all baseline models on these two additional datasets.

\end{document}